\documentclass[journal,10pt]{IEEEtran}  
\usepackage{amsfonts}
\IEEEoverridecommandlockouts
\ifCLASSINFOpdf
\else
\fi
\usepackage{multirow}
\usepackage{booktabs}
\usepackage{makecell}
\usepackage{graphicx}
\usepackage{subfigure}
\usepackage{cite}
\usepackage{color}
\usepackage[pagebackref=false,breaklinks=false,letterpaper=true,colorlinks,bookmarks=false]{hyperref}
\usepackage{amsmath}
\usepackage{amssymb}
\usepackage{array}
\usepackage{bm}
\usepackage{algorithm}
\usepackage{algorithmic}

\usepackage{setspace}
\usepackage{soul}
\usepackage{listings}
\usepackage{color}
\usepackage{float}

\usepackage{etoolbox}
\makeatletter
\patchcmd{\@makecaption}
  {\scshape}
  {}
  {}
  {}
\makeatletter
\patchcmd{\@makecaption}
  {\\}
  {.\ }
  {}
  {}
\makeatother

\definecolor{dkgreen}{rgb}{0,0.6,0}
\definecolor{gray}{rgb}{0.5,0.5,0.5}
\definecolor{mauve}{rgb}{0.58,0,0.82}

\usepackage{cite}
\usepackage{lettrine} 

\title{One-to-Many Semantic Communication Systems: Design, Implementation, Performance Evaluation }
\author{
Han Hu, Xingwu Zhu, Fuhui Zhou, Wei Wu, Rose Qingyang Hu, and Hongbo Zhu
\thanks{
This work was support in part by the National Natural Science Foundation of China under Grant 62222107, Grant 62071223, Grant 62031012, Grant 61871446, and Young Elite Scientist Sponsorship Program by CAST; and in part by Jiangsu Provincial Key Research and Development Program under Grant BE2020084-1; and in part by the National Natural Science Foundation of China under Grant 92067201. The corresponding author is Fuhui Zhou.

H.~Hu, X.~Zhu and H.~Zhu are with the Jiangsu Key Laboratory of Wireless Communications, Nanjing University of Posts and Telecommunications, Nanjing 210003, China, and also with the Engineering Research Center of Health Service System Based on Ubiquitous Wireless Networks, Ministry of Education, Nanjing University of Posts and Telecommunications, Nanjing 210003, China (email: han\_ h@njupt.edu.cn, zhuxw98@163.com, zhuhb@njupt.edu.cn)

F.~Zhou is with College of Electronic and Information Engineering, Nanjing University of Aeronautics and Astronautics, 210000, P. R. China (email: zhoufuhui@ieee.org)

W.~Wu is with the College of Communication and Information Engineering, Nanjing University of Posts and Telecommunications, 210003, P. R. China (email: weiwu@njupt.edu.cn)

R.~Q.~Hu are with the Department of Electrical and Computer Engineering, Utah State University, Logan, UT, USA. (email: rose.hu@usu.edu)
}
}

\begin{document}

\maketitle

\begin{abstract}
Semantic communication in the 6G era has been deemed a promising communication paradigm to break through the bottleneck of traditional communications. However, its applications for the multi-user scenario, especially the broadcasting case, remain under-explored. To effectively exploit the benefits enabled by semantic communication, in this paper, we propose a one-to-many semantic communication system. Specifically, we propose a deep neural network (DNN) enabled semantic communication system called MR\_DeepSC. By leveraging semantic features for different users, a semantic recognizer based on the pre-trained model, i.e., DistilBERT, is built to distinguish different users. Furthermore, the transfer learning is adopted  to speed up the training of new receiver networks. Simulation results demonstrate that the proposed MR\_DeepSC can achieve the best performance in terms of BLEU score than the other benchmarks under different channel conditions, especially in the low signal-to-noise ratio (SNR) regime.   
\end{abstract}


\section{Introduction}
\lettrine[lines=2]{W}{ith}  the rapid development of artificial intelligence (AI) and natural language processing (NLP), intelligent communication is envisioned as a promising solution to unlocking the bottleneck of traditional communication systems\cite{Qin2019deep}. Especially, deep learning-based semantic communications have shown great potential in realizing the next level of communication, and have attracted widespread attention lately. Compared with traditional communication systems, deep learning-based semantic communication systems only transmit the basic semantic information at the transmitter, and reconstruct the semantic information through prior knowledge at the receiver, which can significantly reduce the required communication resources and achieve robust performance in the bad channel environments, i.e., low signal-to-noise (SNR) ratio regime\cite{lan2021semantic}.

Due to the great potential of deep learning-enabled semantic communication, many researchers have focused on the system design for various source contents \cite{n2018deep,xie2021deep,zhou2021cognitive,weng2020se,Bourtsoulatze2019deep,Kurka2020deepjcss}. The authors in \cite{n2018deep} proposed a joint source-channel coding scheme to transmit text sentences with fixed length in simple channel environments. In order to handle text sentences with different lengths more flexibly in complex channel environments, the authors in \cite{xie2021deep} further developed a Transformer-based semantic communication framework. Moreover, a semantic communication system combined with a knowledge graph was developed in \cite{zhou2021cognitive} to further improve semantic error correction capability and data compression rate. In addition to text transmission, an attention-based semantic communication was designed to process speech signals in \cite{weng2020se}. Recently, the works  \cite{Bourtsoulatze2019deep} and \cite{Kurka2020deepjcss}  investigated the deep learning-enabled semantic communication for image transmission. 

It is worth noting that all of the works mentioned above mainly focused on single-user semantic communication systems. In reality, with the emergence of new applications, such as autonomous transportation, drone fleets, and remote command systems, the multi-user system for semantic transmission needs to be explored to support various user requirements. The latest work\cite{xie2021mu} designed a task-oriented semantic communication system for multi-user cases, wherein many-to-one and many-to-many communication for different tasks were investigated. However, the broadcast case with one transmitter and multiple receivers is not considered, which is also very important in wireless communications.

Motivated by the research efforts mentioned above, in this paper, we aim to develop a deep learning-based semantic communication system for the one-to-many broadcast scenario. Firstly, a novel semantic communication framework consisting of one transmitter and multiple receivers based on Transformer is developed.  
Secondly, considering that different users possess different semantic information,  the pre-trained model, i.e., DistilBERT, is built as the semantic recognizer at each receiver to distinguish users. 
Furthermore, in order to deal with multiple different channel environments experienced by different users,  the deep transfer learning is adopted to speed up the training processes of the new receiver network.
Finally, simulation results demonstrate that the proposed framework is superior to the traditional communication model and some other DL-based semantic models in terms of BLEU score and has a robust performance in various channel environments, especially in the case of a low SNR ratio.

The rest of this paper is organized as follows.  Section II describes the system model for one-to-many semantic communication. In Section III, the proposed MR\_DeepSC system model is detailed. Section IV presents simulation results to evaluate the performance of MR\_DeepSC. Finally, Section V provides the conclusion.

\section{One-to-Many Semantic Communication System Design}
The proposed system model is shown in Fig. \ref{fig:framework}, where we consider a DNN-enabled semantic communication model for one-to-many communication that consists of a transmitter and multiple receivers. The transmitter is responsible for transmitting all the users' source text information while each receiver only estimates its own information.
In this regard, the proposed system is expected to accomplish the following major tasks:
i) recover the original information as accurately as possible; 
ii) distinguish different users' information at receivers. 
\subsection{Transmitter}
As illustrated in Fig. \ref{fig:framework}, the source sentence of  user $k$ is represented as ${{\bf{s}}_{\bf{k}}} = [w_1^k,w_2^k,...,w_{{L_k}}^k]$, where $w_l^k$ represents the $l$-th word in the sentence of user $k$.
We consider that the transmitter packets all source sentences without knowing the corresponding user of each sentence,
so all source sentences are shuffled and merged into a long sequence as the input ${\bf{s}} = [w_1^k,w_2^k,...,\langle sep\rangle ,w_1^1,w_2^1,...]$,  where $\langle sep\rangle $ is the separator between each sentence.
The transmitter consists of two components, namely the semantic encoder and the channel encoder, where the semantic information from the transmitted sequence $s$ is extracted by the semantic encoder and then is transmitted over physical channels after channel coding by the channel encoder. Specifically, the semantic encoder first calculates the dependencies among words in different positions of the sentence. Then, it extracts the important semantic information according to the importance of the dependencies.
The channel encoder plays channel encoding on the extracted semantic information for transmission on the physical channel. Note that  semantic encoder and channel encoder are implemented by  independent neural networks respectively. 
 Thus, the encoded symbol sequence  ${\bf{x}}\in {\mathbb{C}}^{M\times 1}$ is written  as
\begin{equation}
\setlength{\abovedisplayskip}{3pt}
\setlength{\belowdisplayskip}{3pt}
{\bf{x}} = {T^C}\left( {{T^S}\left( {{\bf{s}};\alpha } \right);\beta } \right),
\end{equation}
where $M$ is the length of the symbol sequence, 
${T^S}\left( { \cdot \;;\alpha } \right)$  represents the semantic encoder constructed based on  deep neural networks, and $\alpha $ is the parameter set of the deep neural network.
${T^C}\left( { \cdot \;;\beta } \right)$ represents the channel encoder, which is constructed by  the neural networks with parameter set $\beta$.

\subsection{Receivers}
\begin{figure}[t]
	\includegraphics[width=1.00\linewidth]{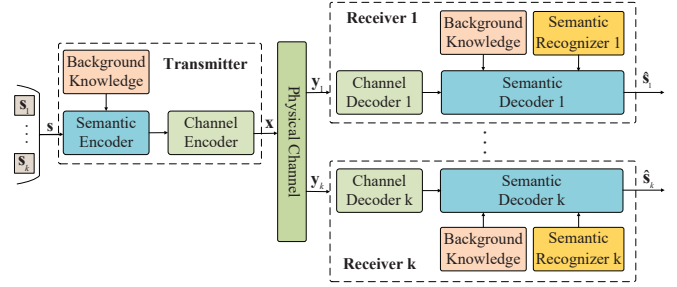}
	\vspace{-0.25in}
	\caption{The framework of one-to-many semantic communication system. }
	\label{fig:framework}
	\vspace{-0.25in}
\end{figure}

Different from the point-to-point communication transmission for a single user,  the communication system designed for broadcast communications in Fig. \ref{fig:framework} involves one transmitter and multiple receivers. When the transmitted signal transverses physical channels, the received signal ${\bf{y}}_k\in {\mathbb{C}}^{M\times 1}$ at receiver $k$ is expressed as
\begin{equation}
\setlength{\abovedisplayskip}{3pt}
\setlength{\belowdisplayskip}{3pt}
{{\bf{y}}_{k}} = {h_k}{\bf{x}} + {{\bf{w}}_{k}},
\end{equation}
where ${h_k}$ represents the coefficients of the linear channel, ${{\bf{w}}_{{k}}} \sim \mathcal{CN}\left( {0, \delta _n^2} \right)$ indicates independent and identically distributed Gaussian noise.

As shown in Fig. \ref{fig:framework}, there are also two main parts in each receiver structure, i.e., channel decoder and semantic decoder. The channel decoder is used to recover the transmitted symbols while the semantic decoder is used to recover the transmitted sentences.
The decoded signal of receiver $k$ can be formulated  as
\begin{equation}
\setlength{\abovedisplayskip}{3pt}
\setlength{\belowdisplayskip}{3pt}
{{\bf{\hat s}}_{{k}}} = R_k^S\left( {R_k^C\left( {{{\bf{y}}_{{k}}};{\chi _k}} \right);{\delta _k}} \right),
\end{equation}
where ${{\bf{\hat s}}_{{k}}}$ represents the target sentence of user $k$, $R_k^C\left( {\cdot \;;{\chi _k}} \right)$ is the channel decoder of receiver $k$ with the parameter set ${\chi _k}$,  and $R_k^S\left( {\cdot \;;{\delta _k}} \right)$ represents the semantic decoder of receiver $k$ with the parameter set ${\delta _k}$.

Note that in the broadcast case, the transmitted signals and sentences include information from all users. In order to improve resource efficiency, the traditional division mechanisms (e.g., TDMA or FDMA) are not utilized in our design.
Considering the sentences sent to different users may carry different semantic features, such as language, emotions, etc,  we establish a semantic recognizer at each receiver to label these differences. 
Combing the background knowledge and the semantic recognizer, only the received information satisfying the target user's feature is finally received at each receiver.

\section{One-to-Many Semantic Communication system Implementation }
 In this section, without loss of generality,  we consider two users and distinguish them according to the different emotions for simplicity, i.e., user 1 receives positive text sentence ${S^P}$ whereas user 2 receives negative text sentence ${S^N}$. 
 Then, a DNN-enabled semantic communication system named  MR\_DeepSC is proposed, in which Transformer \cite{vaswani2017attention} is exploited for semantic extraction and recovery and DistilBERT\cite{VSanh2019DistilBERT} is adopted as a semantic recognizer.
 
 \begin{figure*}[t]
 \setlength{\abovedisplayskip}{3pt}
\setlength{\belowdisplayskip}{3pt}
\vspace{-0.3cm}
	\setlength{\abovecaptionskip}{-0.2cm} 
	\setlength{\belowcaptionskip}{-1cm}
	\centering
\centering
\includegraphics[width=0.85\linewidth]{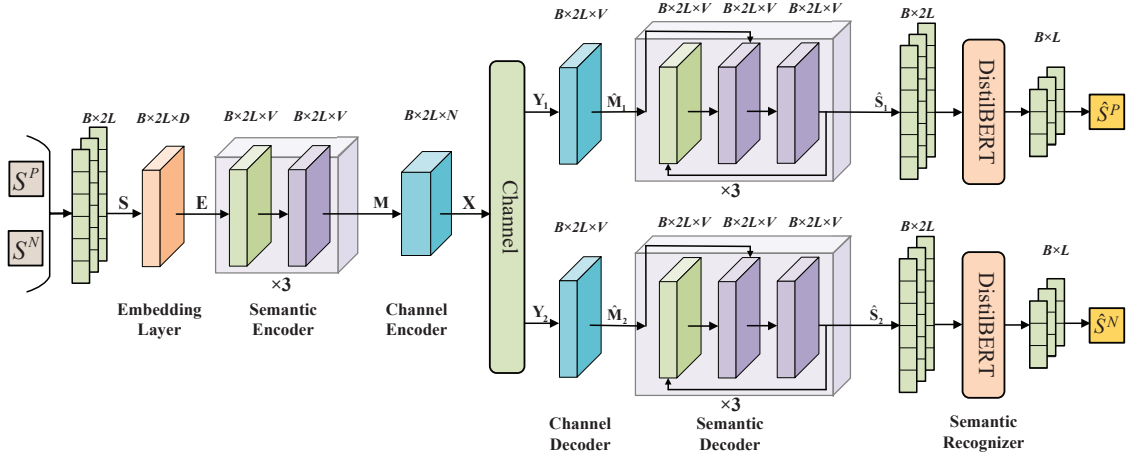}
\caption{The proposed system architecture for the one-to-many semantic communication system}
\label{fig:model}
\vspace{-0.25in}
\end{figure*}
\subsection{Model Description}
As shown in Fig. \ref{fig:model}, a small batch of input sentences ${\bf{S}} \in {\Re ^{B \times 2L}}$ is generated by a knowledge set $\mathcal{D}$, where $B$ is the batch size. Each sentence consists of one positive sentence ${S^P}$ and one negative sentence ${S^N}$, both of which are padded to the same length $L$ by special symbols.
The embedding layer converts the words of the sentence into word vectors and obtains the word vector sequence ${\bf{E}} \in {\Re ^{B \times 2L \times D}}$ as the input of the semantic encoder. Here $D$ is the dimension of each word vector. The semantic encoder consists of multiple Transformer encoding layers, each of which is further divided into two sublayers, i.e, the self-attention sublayer and the feed-forward sublayer. The self-attention layer is first explored to transform the current input  ${\bf{E}}$ into three matrix, i.e., the query  matrix ${\bf{Q}} \in {\Re ^{B \times 2L \times V}}$, the key matrix ${\bf{K}} \in {\Re ^{B \times 2L \times V}}$, and the value matrix ${\bf{V}} \in {\Re ^{B \times 2L \times V}}$ through three different linear layers. $V$ is the output dimension of the three linear layers. 
Then the attention operation is performed based on these three matrices to obtain the matrix ${\bf{M}} \in {\Re ^{B \times 2L \times V}}$ as the semantic information.
The feed-forward sublayer is composed of two linear layers to improve the fitting degree of the network. Subsequently, the dense layers of the channel encoder convert these semantic information ${\bf{M}}$ into transmitted symbols ${\bf{X}} \in {\Re ^{B \times 2L \times N}}$ suitable for channel transmission.

The physical channel layer takes $\bf{X}$ as input and ${{\bf{Y}}_1}$ and ${{\bf{Y}}_2}$ as outputs at receiver 1 and  receiver 2, respectively. ${{\bf{Y}}_1}$ and ${{\bf{Y}}_2}$ are respectively given as 
\begin{equation}
\setlength{\abovedisplayskip}{3pt}
\setlength{\belowdisplayskip}{3pt}
\begin{array}{l}
{{\bf{Y}}_1} = {{\bf{H}}_1}{{\bf{X}}} + {{\bf{W}}_1}\\
{{\bf{Y}}_2} = {{\bf{H}}_2}{{\bf{X}}} + {{\bf{W}}_2},
\end{array}
\end{equation}
where ${{\bf{H}}_1}$ and ${{\bf{H}}_2}$ contain $B$ vectors of channel coefficient, ${{\bf{W}}_1}$ and ${{\bf{W}}_2}$  consist of $B$ vectors of  Gaussian noise.

Considering that all the receivers in the considered model have the same structure, we take receiver 1 as an example for simplicity. After receiving 
${\bf{Y}}_1$, the dense layers of the channel decoder in receiver 1 reverse  the received symbol sequence to recover the semantic matrix ${{{\bf{\hat M}}}_1}\in {\Re ^{B \times 2L \times V}}$.
The semantic decoder consists of multiple Transformer decoding layers, and each Transformer decoding layer has three sublayers, i.e., the self-attention sublayer, encoder-decoder attention sublayer, and feed-forward sublayer.
The self-attention sublayer performs the attention operation on the past output to obtain the query matrix. The encoder-decoder attention sublayer passes the semantic matrix ${{{\bf{\hat M}}}_1}$ through different linear layers to obtain the key matrix and the value matrix, and then performs the attention operation based on these three matrices and estimates the original sentence ${{{\bf{\hat S}}}_1}$.

Since ${{{\bf{\hat S}}}_1}$ contains sentences of different users, we separate sentences with equal length first and then apply the pre-trained model DistilBERT to categorize sentence emotions of these sentences belonging to different users.
As a compressed version of BERT, DistilBERT is smaller, faster, and lighter than the typical BERT, and has been trained by millions of sentences, which makes it ready for a variety of tasks \cite{VSanh2019DistilBERT}. 
For each sentence, DistilBERT utilizes multiple Transformer encoding layers to output a vector for representing its global information, which is then further input into the classification layer to obtain the emotional features of each sentence.
Finally, the positive sentence required by the user 1 can be extracted from ${{{\bf{\hat S}}}_1}$ according to these emotional features.

\subsection{Training Algorithm}
The training process of the whole system is illustrated 
in Fig. \ref{fig:train} and the pseudocode is given in Algorithm \ref{alg:e2e}. 
The whole training process is  divided into two phases. In  the first phase, it aims to train the network between the transmitter and receiver 1. 
Although receiver 1 and receiver 2 have a similar network structure and share one transmitter, since different users have different transmit channels, in the second phase, we adopt the deep transfer learning to train receiver 2 to reduce the training cost and improve the training speed.

Specifically, in the first phase,  a small batch of input $\bf{S}$ from the knowledge set $\mathcal{D}$ is encoded into $\bf{M}$ through a semantic encoder. Then,  $\bf{M}$ is converted into $\bf{X}$ by channel encoder over the physical channel. At receiver 1,  ${{\bf{Y}}_1}$ is received and then decoded at the physical channel layer to obtain the recovered semantic information  ${{{\bf{\hat M}}}}$. Afterwards, the  semantic decoder layer is utilized to estimate the semantic sentence ${{{\bf{\hat S}}}_1}$. Note that ${{{\bf{\hat S}}}_1}$ is not processed by the semantic recognizer and contains all the semantic sentences from the input. 
Finally, the network between the transmitter and receiver 1 is trained by the stochastic gradient descent with the cross-entropy loss function ${{\cal L}_{CE}}({\bf{S}}, {{{\bf{\hat S}}}_1}; \alpha, \beta, {\chi _1}, {\delta _1})$ as follows.
\begin{equation}
\setlength{\abovedisplayskip}{3pt}
\setlength{\belowdisplayskip}{3pt}
	\setlength{\abovecaptionskip}{-0.2cm} 
	\setlength{\belowcaptionskip}{-1cm}
\begin{array}{l}
{{\cal L}_{CE}}({\bf{S}}, {{{\bf{\hat S}}}_1}; \alpha, \beta, {\chi _1}, {\delta _1}) = \\
 - \sum\limits_{l = 1} {\left( {q\left( {{w_l}} \right)\log \left( {p\left( {{w_l}} \right)} \right) + \left( {1 - q\left( {{w_l}} \right)} \right)\log \left( {1 - p\left( {{w_l}} \right)} \right)} \right)},
\end{array}
\end{equation}
where ${q\left( {{w_l}} \right)}$ and ${p\left( {{w_l}} \right)}$ represent the actual and predicted  probabilities of the $l$-th word appearing in ${\bf{S}}$ and ${{{\bf{\hat S}}}_1}$, respectively.

In the second phase, we first load the pre-trained transmitter and receiver 1. For a different receiver, we only need to redesign and train the semantic decoder and channel decoder after freezing the parameters of the semantic and channel encoder in the same transmitter. Then we repeat the steps of the first phase to train receiver 2 until convergence.

\begin{algorithm}[t]
\begin{spacing}{1.15}
\newcommand{\INDSTATE}[1][1]{\STATE\hspace{#1\algorithmicindent}}
\caption{MR\_DeepSC Training Algorithm.} 
\label{alg:e2e} 
\begin{algorithmic}[1] 
\REQUIRE The background knowledge set $\mathcal{D}$;
\ENSURE \textit{Train Transmitter and Receiver 1}:
\STATE \textbf{Initial the weights W and bias b};
\INDSTATE[0]\textbf{Transmitter:}
\INDSTATE[2] Take a batch \bm{$S$} from the set $\mathcal{D}$
\INDSTATE[2] ${T^S}\left( {\bf{S};\alpha } \right) \to \bf{M}$.
\INDSTATE[2] ${T^C}\left( {\bf{M};\beta } \right) \to \bf{X}$.
\INDSTATE[2] Transmit $\bf{X}$ over the channel.
\INDSTATE[0]\textbf{Receiver 1:}
\INDSTATE[2] Receive ${{\bf{Y}}_1}$.
\INDSTATE[2] $R_1^C\left( {{{\bf{Y}}_1};{\chi _1}} \right) \to {{{\bf{\hat M}}}_1}$.
\INDSTATE[2] $R_1^S\left( {{{{\bf{\hat M}}}_1};{\delta _1}} \right) \to {{{\bf{\hat S}}}_1}$
\INDSTATE[2] Compute loss function ${{\cal L}_{CE}}({\bf{S}}, {{{\bf{\hat S}}}_1}; \alpha, \beta, {\chi _1}, {\delta _1})$.
\INDSTATE[2] Train $\alpha, \beta, {\chi _1}, {\delta _1}$ $ \to $ Gradient descent with ${{\cal L}_{CE}}$.

\STATE \textbf{Return: } ${T^S}\left( {\cdot \;;\alpha } \right)$, ${T^C}\left( {\cdot \;;\beta } \right)$, $R_1^C\left( {\cdot \;;{\chi _1}} \right)$, $R_1^S\left( {\cdot \;;{\delta _1}} \right)$.

\ENSURE \textit{Transfer learning based training for Receiver 2}:

\STATE \textbf{Load the pre-trained model: } ${T^S}\left( {\cdot \;;\alpha } \right)$, ${T^C}\left( {\cdot \;;\beta } \right)$, $R_1^C\left( {\cdot \;;{\chi _1}} \right)$, $R_1^S\left( {\cdot \;;{\delta _1}} \right)$.
\STATE Freeze ${T^S}\left( {\cdot \;;\alpha } \right)$ and ${T^C}\left( {\cdot \;;\beta } \right)$.
\STATE ${\chi _1} \to {\chi _2}, {\delta _1} \to {\delta _2}$.
\STATE Repeat line 2-12 to train Receiver 2  until convergence.
\STATE \textbf{Return: } $R_2^C\left( {\cdot \;;{\chi _2}} \right)$, $R_2^S\left( {\cdot \;;{\delta _2}} \right)$.

\end{algorithmic}
\end{spacing} 
\end{algorithm}

\section{Performance Evaluation}
In this section, we adopt the widely recognized evaluation metric in natural language processing, namely the bilingual evaluation understudy (BLEU)\cite{papineni2002bleu}, for measuring the performance of different approaches, and  compare the proposed MR\_DeepSC model with the other four benchmarks under both the AWGN channel and Rayleigh fading channel.

\begin{figure}[t]
\setlength{\abovedisplayskip}{3pt}
\setlength{\belowdisplayskip}{3pt}
	\setlength{\abovecaptionskip}{-0.2cm} 
	\setlength{\belowcaptionskip}{-1cm}
    \centering
	\includegraphics[width=0.7\linewidth]{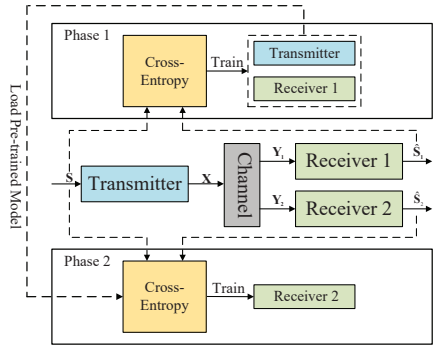}
	\caption{The training framework of the MR\_DeepSC. }
	\label{fig:train}
	\vspace{-0.2in}
\end{figure}

\subsection{Simulation Settings}

\begin{figure*}[h]
\centering
\begin{minipage}[h]{0.62\linewidth}
\subfigure[]{\includegraphics[width=5.5cm]{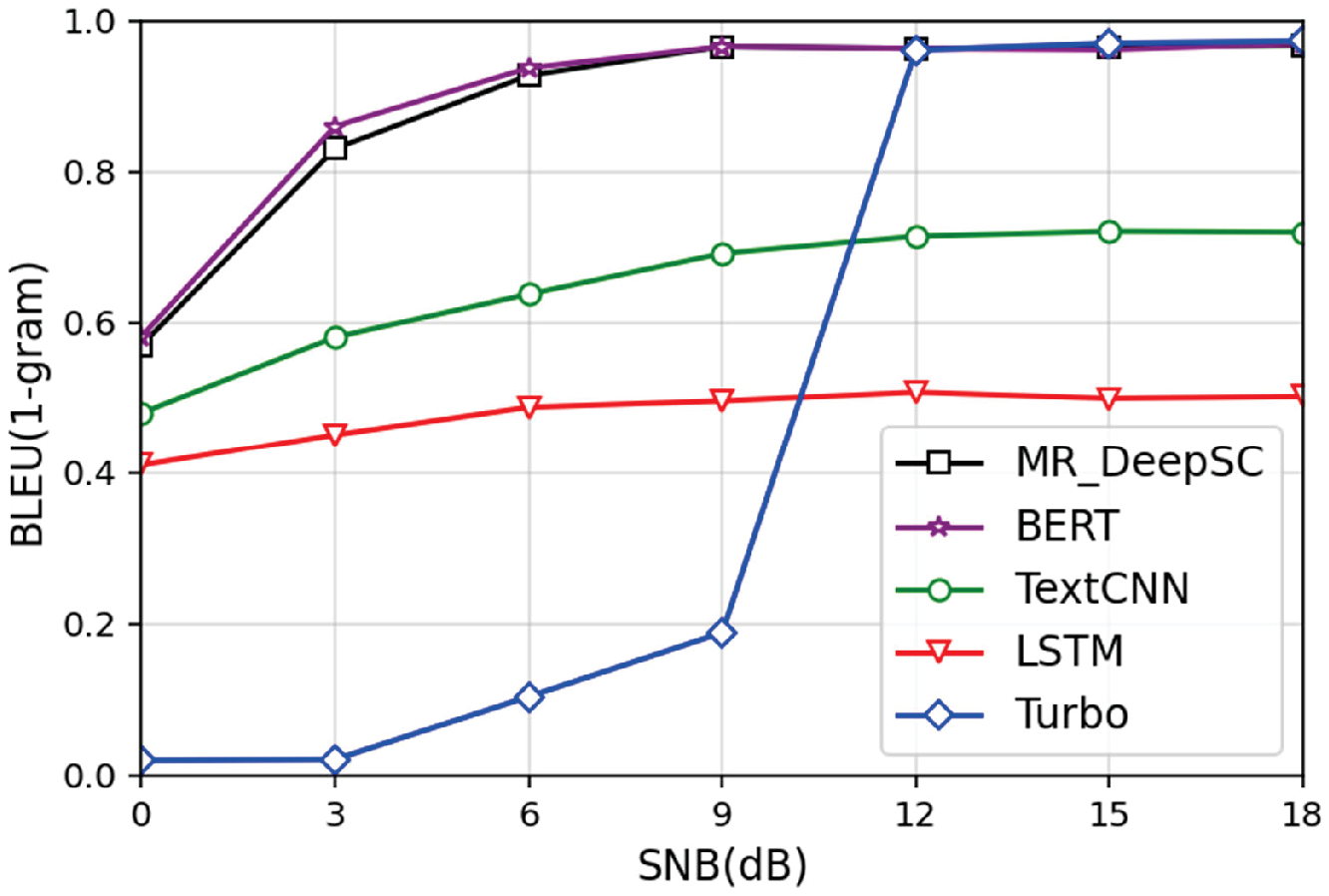}\label{fig:awgn}}
\subfigure[]{\includegraphics[width=5.5cm]{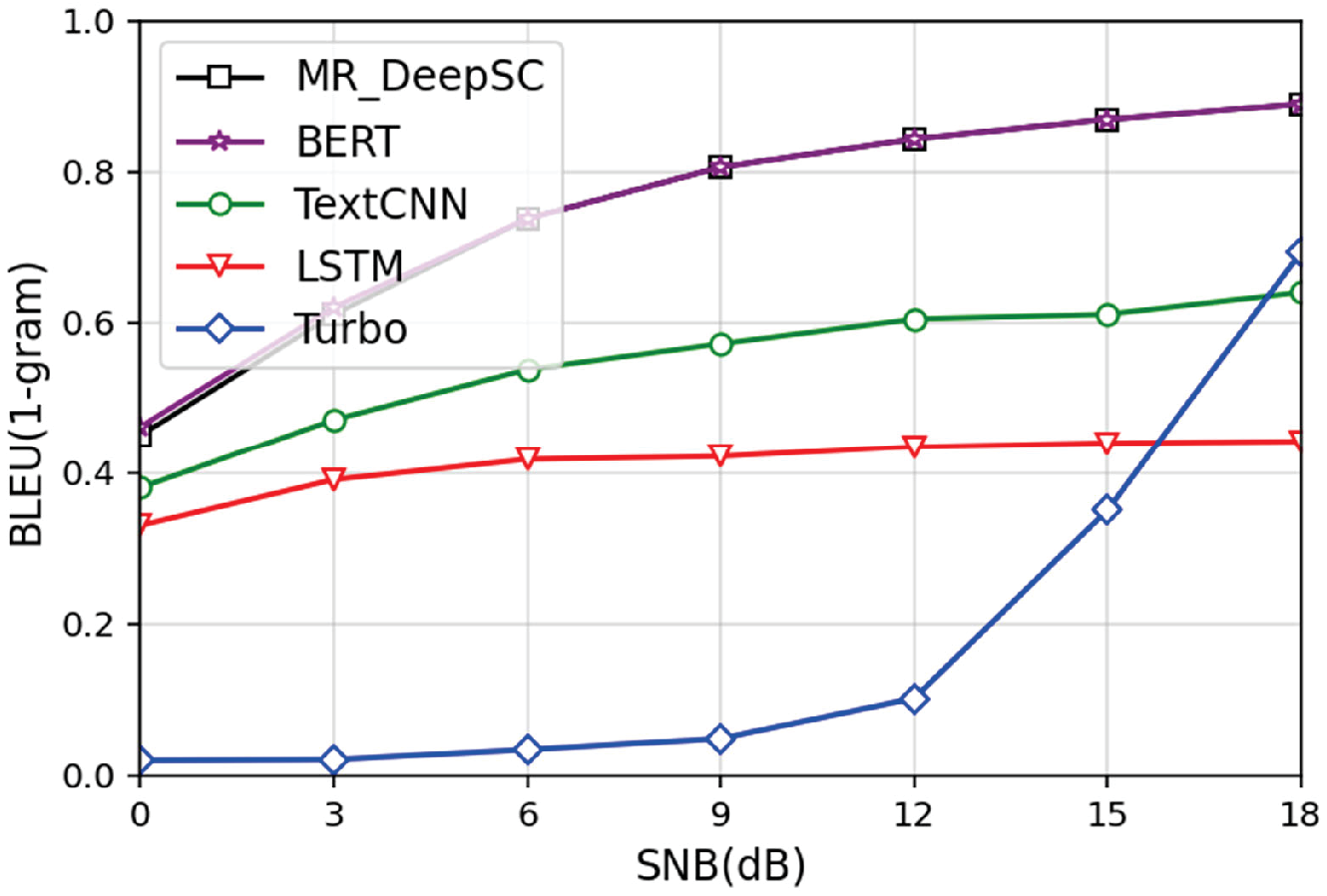}\label{fig:rey}}
\vspace{-0.15in}
\caption{(a) BLEU score of receiver 1 tested under AWGN channels;
(b) BLEU score of receiver 2 tested under Rayleigh Fading channels.}
\end{minipage}
\begin{minipage}[h]{0.3\linewidth}
\centering
\includegraphics[width=5.5cm]{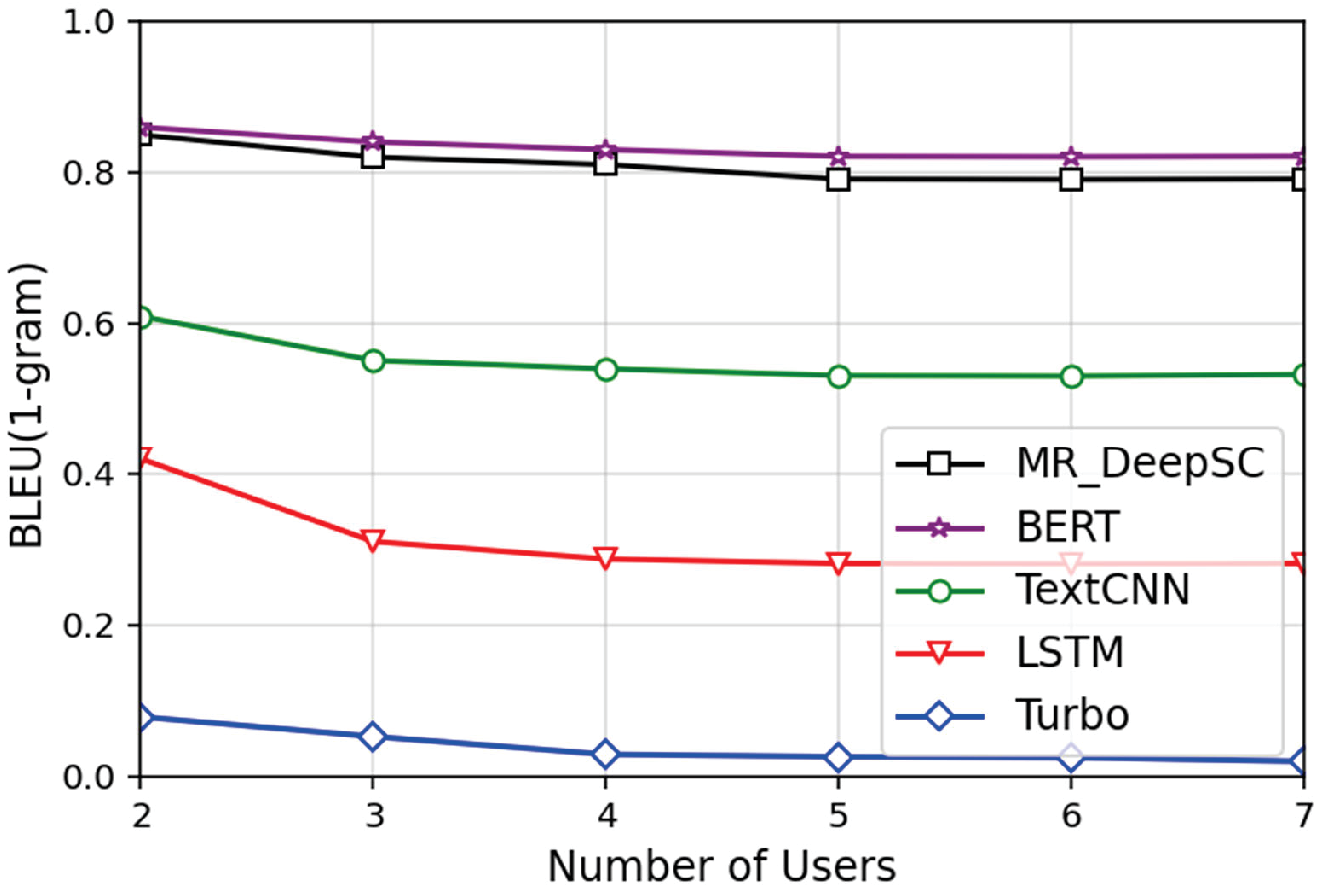}
\vspace{-0.12in}
\caption{BLEU score comparisons versus the number of users.}
\label{fig:user}
\end{minipage}
\label{fig:result}
\vspace{-0.2in}
\end{figure*}

Three datasets for training and testing are adopted in our numerical simulation. The first dataset is the standard proceedings of the European Parliament\cite{Koehn2005Europarl}, which consists of around $2$ million sentences and is pre-processed into sentences of $4$ to $15$ words. The second comes from the Internet including $8000$ positive sentences and $8000$ negative sentences. As the mechanism to distinguish users is based on their obvious emotional features, we divide the second dataset into two parts. The first part along with the first dataset is used as the training dataset while  the  remaining part is used as the testing dataset.
The third dataset comes from the Internet, which contains seven categories including sports, education, finance, games, medicine, politics, and military.
Each category contains 6000 sentences with distinct distinguishing features. It is divided in the same way as the second dataset and used for the system with more than two users in Fig. \ref{fig:user}.
The typical Transformer that consists of three encoding and decoding layers with $8$ heads is utilized for the semantic coding and decoding. The channel encoder and decoder are set as a dense layer of $16$ units and $128$ units, respectively. A single antenna is adopted for each transmitter and receiver. The whole network is optimized by the SGD and the learning rate is $1 \times {10^{ - 4}}$.
For performance comparison, we provide the other four benchmarks which are defined as follows.

1) BERT: A DNN-based communication system in which the structure of semantic encoder and semantic decoder is the same as that of the proposed MR\_DeepSC, but the semantic recognizer is constructed based on the pre-trained model BERT \cite{VSanh2019DistilBERT}.

2) TextCNN: A DNN-based semantic communication system similar to the proposed method except that the TextCNN-based semantic recognizer  \cite{Y2014TextCNN} is adopted at the receiver and is pre-trained with the IMDb dataset \cite{maas2011learning}.


3) LSTM: A DNN-based communication system in which the semantic encoder and semantic decoder are reconstructed based on the LSTM in \cite{n2018deep}, and a DistilBERT-based semantic recognizer is exploited to distinguish different users.
 
4) Turbo: A traditional communication system in which the source coding and channel coding are employed independently. The source coding is Huffman coding while the channel coding is Turbo coding\cite{CHeegard2013Turbo}. The turbo encoding rate is $1/3$ and the Max-Log-MAP algorithm with $5$ iterations is used for the turbo decoding. Moreover, $64$-QAM and CDMA are exploited for the modulation and  multiple access, respectively.

\subsection{Simulation Results}
Fig. \ref{fig:awgn} and Fig. \ref{fig:rey} illustrate the relationship between BLEU score and SNR value for two receivers of different benchmarks under different channel environments, where receiver 1 and receiver 2 are tested under the AWGN channel and the Rayleigh fading channel, respectively.
In the comparison between the DNN-based models and the traditional communication model, it is observed that although the traditional communication model ``Turbo" has better performance in the case of high SNR, the performance and stability of all the DNN-based models in the low SNR domain are significantly better than the traditional communication model due to the only basic semantic information at the transmitter in the semantic communication systems.
Moreover, all DNN-based semantic communication systems are more robust to different channel conditions, especially in the low SNR domain.
The performance of our proposed ``MR\_DeepSC'' is better than the ``LSTM"  and the ``TextCNN'' but is similar to the  ``BERT" method. 
The reason can be explained as follows: since the structure of the LSTM layer is relatively simple than the DNN-based structure utilized in the other three DNN-based methods, it always makes the model unable to correctly capture the dependencies between two distant words \cite{n2018deep}. 
The  TextCNN-based  recognizer of the ``TextCNN'' is trained by a limited dataset while the DistilBERT is pre-trained with billions of data, thus the proposed ``MR\_DeepSC" method is able to achieve a higher BLEU score than the ``TextCNN''.
Despite the ``BERT " method performing similarly to the proposed ``MR\_DeepSC",  it will cost more computing resources to maintain better performance due to a large number of parameters of the BERT recognizer\cite{VSanh2019DistilBERT}.

\begin{table}[t]
\scriptsize
\centering  
\setlength{\abovecaptionskip}{0cm}
\caption{Comparison of computational complexity }
\renewcommand\arraystretch{1.3}
\label{Table1} 
\setlength{\tabcolsep}{1mm}{
\begin{tabular}{|c|cc|cc|}
\hline
\multirow{2}{*}{Methods} & \multicolumn{2}{c|}{Complexity of Encoder-Decoder} & \multicolumn{2}{c|}{Complexity of  Recognizer}   \\ \cline{2-5} 
                         & \multicolumn{1}{c|}{Additions}  & Multiplications  & \multicolumn{1}{c|}{Additions} & Multiplications \\ \hline
MR DeepSC                & \multicolumn{1}{c|}{$3.4 \times {10^5}$}        & {$3.4 \times {10^5}$}              & \multicolumn{1}{c|}{$6.3 \times {10^7}$}       & {$6.3 \times {10^7}$}             \\ \hline
LSTM                     & \multicolumn{1}{c|}{$2.3 \times {10^5}$}        & {$2.3 \times {10^5}$}              & \multicolumn{1}{c|}{$6.3 \times {10^7}$}       & {$6.3 \times {10^7}$}             \\ \hline
BERT                     & \multicolumn{1}{c|}{$3.4 \times {10^5}$}        & {$3.4 \times {10^5}$}              & \multicolumn{1}{c|}{$11.2 \times {10^7}$}      & {$11.2 \times {10^7}$}             \\ \hline
TextCNN                  & \multicolumn{1}{c|}{$3.4 \times {10^5}$}        & {$3.4 \times {10^5}$}             & \multicolumn{1}{c|}{$2.1 \times {10^5}$}       & {$2.1 \times {10^5}$}              \\ \hline
Turbo                    & \multicolumn{1}{c|}{$1.2 \times {10^5}$}        & {$1.2 \times {10^5}$}              & \multicolumn{2}{c|}{None}                          \\ \hline
\end{tabular}
}
\vspace{-0.2in}
\end{table}

The computational complexity for different models is compared in Table \ref{Table1}.
The traditional model ``Turbo"  has a low complexity than all DNN-based models due to the complicated structure of deep neural networks. The complexity of ``LSTM'' and ``TextCNN'' is lower than that of the proposed model since the structures of the LSTM-based encoder-decoder in ``LSTM''  and the TextCNN-based recognizer in ``TextCNN''  are relatively simple.
Besides, the proposed model shows a much lower complexity compared with the ``BERT" method for the reason that the DistilBERT adopted in the proposed model is a compressed version of the BERT,  which has much fewer structural layers. As a result, we can conclude that our proposed method can achieve a better balance between complexity and performance.

Fig. \ref{fig:user} illustrates the impact of the number of users  on the performance with different models in Rayleigh fading channels with an SNR of $12$ dB. The performance metric is the average of all users' BLEU scores. 
When the number of users increases, longer input sentences cause more difficulty for the encoder, resulting in a slight decrease in the performance of all models.
In addition, it demonstrates that the proposed model achieves better performance than the other four benchmarks.

\section{Conclusion}
In this paper, we proposed a DNN-enabled semantic communication system called ``MR\_DeepSC" for one-to-many communications. The semantic coding and channel coding were jointly designed to learn and extract the features in order to achieve robust  performance under various channel conditions.
The transfer learning was adopted to speed up the training of the new receiver network. 
By taking advantage of different emotional features, a semantic recognizer based on the pre-trained model was developed to distinguish different users. 
Simulation results demonstrated that the proposed system can improve the performance gains compared with other benchmarks under different channel conditions.
To the best of the authors' knowledge, the work in this paper is the first attempt to design a one-to-many semantic communication system. The proposed ``MR\_DeepSC" can pave the way for the development of future semantic communication systems.

\bibliographystyle{IEEEtran}

\begin{thebibliography}{10}

\bibitem{Qin2019deep}
Z.~Qin \emph{et~al.}, ``Deep learning in physical layer communications,'' \emph{IEEE Wireless Commun.}, vol.~26, no.~2, pp. 93--99, 2019.

\bibitem{lan2021semantic}
Q.~Lan \emph{et~al.}, ``What is semantic communication? A view on conveying meaning in the era of machine intelligence,''
  \emph{Journal of Communications and Information Networks.}, vol.~6, no.~4, pp. 336--371, 2021.

\bibitem{n2018deep}
N.~Farsad \emph{et~al.}, ``Deep learning for joint sourcechannel
coding of text,''
  \emph{IEEE Int’l. Conf. Acoustics Speech
Signal Process. (ICASSP).}, algary, AB, Canada,  pp. 2326--2330, 2018.

\bibitem{xie2021deep}
H.~Xie \emph{et~al.}, ``Deep learning enabled
semantic communication systems,''
  \emph{IEEE Trans. Signal Process.}, vol.~69, pp. 2663--2675, 2021.


\bibitem{zhou2021cognitive}
F.~Zhou \emph{et~al.}, ``Cognitive semantic communication systems driven by knowledge graph,'' \emph{IEEE ICC.}, to be published, 2022.
  
\bibitem{weng2020se}
Z.~Weng \emph{et~al.}, ``Semantic communication systems for speech
transmission,'' \emph{IEEE J. Sel. Areas Commun.}, vol.~39, no.~8, pp. 2434--2444, 2021.

\bibitem{Bourtsoulatze2019deep}
E.~Bourtsoulatze \emph{et~al.}, ``Deep joint sourcechannel
coding for wireless image transmission,'' \emph{IEEE Trans. Cogn.
Commun. Netw.}, vol.~5,
  no.~3, pp. 567--579, 2019.

\bibitem{Kurka2020deepjcss}  
D. B.~Kurka \emph{et~al.}, ``DeepJSCC-f: Deep joint source-channel
coding of images with feedback,'' \emph{IEEE J. Select. Areas Inf. Theory.}, vol.~1, no.~1, pp. 178–-193, 2020.
  
\bibitem{xie2021mu}
H.~Xie \emph{et~al.}, ``Task-oriented multi-user semantic communications,'' in \emph{arXiv preprint arXiv:2112.10255}, 2021.

\bibitem{vaswani2017attention}
A.~Vaswani \emph{et~al.}, ``Attention is all you need,'' in \emph{Adv. neural inf.proces. syst.}, 2017, pp. 5999--6009.
  
\bibitem{VSanh2019DistilBERT}
V.~Sanh \emph{et~al.}, ``DistilBERT, a distilled version of BERT: smaller, faster, cheaper and lighter,'' in \emph{arXiv preprint arXiv:1910.01108}, 2019.

\bibitem{papineni2002bleu}
K.~Papineni \emph{et~al.}, ``Bleu: a method for automatic evaluation of machine translation,'' in \emph{Proceedings of the 40th annual meeting of the Association for Computational Linguistics}, pp. 311--318, 2019.

\bibitem{Koehn2005Europarl}
P.~Koehn \emph{et~al.}, ``Europarl: A parallel corpus for statistical machine translation,'' \emph{MT summit.}, vol.~5, Citeseer, pp. 79--86, 2005.

\bibitem{Y2014TextCNN}
Y.~Kim \emph{et~al.}, ``Convolutional Neural Networks for Sentence Classification,'' in \emph{arXiv preprint 	arXiv:1408.5882}, 2014.

\bibitem{maas2011learning}
A.~Maas \emph{et~al.}, ``Learning word vectors for sentiment analysis.,'' \emph{Proceed in} Association for Computational Linguistics (ACL), pp. 142--150, 2011.

\bibitem{CHeegard2013Turbo}
C.~Heegard \emph{et~al.}, ``Turbo coding,'' in \emph{Springer Science \& Business Media}, Vol.~476, 2013.

\end{thebibliography}

\end{document}